\documentclass{fairmeta}

\usepackage[utf8]{inputenc}
\usepackage{amsmath,amssymb}
\usepackage{array,colortbl,adjustbox}
\usepackage{algorithm}
\usepackage{algpseudocode}
\usepackage{xurl,wrapfig,needspace,float}
\usepackage{fontawesome5}

\usepackage{amsmath,amsfonts,bm}

\def\eqref#1{equation~\ref{#1}}

\def\1{\bm{1}}

\DeclareMathAlphabet{\mathsfit}{\encodingdefault}{\sfdefault}{m}{sl}
\SetMathAlphabet{\mathsfit}{bold}{\encodingdefault}{\sfdefault}{bx}{n}

\titlespacing*{\section}{0pt}{2.5ex plus 0.5ex minus 0.2ex}{1.2ex plus 0.2ex}
\titlespacing*{\subsection}{0pt}{2ex plus 0.5ex minus 0.2ex}{0.8ex plus 0.2ex}
\setlabdisplayname{OmniAI Group of ZJU ACES Lab}
\setuniversityname{Zhejiang University}

\title{DTCA: Denoising Token Credit Assignment for Reinforcement Learning in Diffusion Language Models}
\author[1]{\mbox{Hongze Chen}}
\author[1]{\mbox{Miao Pan}}
\author[1]{\mbox{Lizhou Liang}}
\author[1]{\mbox{Qinfeng Li}}
\author[2]{\mbox{Xinyin Ma}}
\author[1]{\mbox{Yongliang Shen}}
\author[3]{\mbox{Weishi Wang}}
\author[3]{\mbox{Daniel Dahlmeier}}
\author[1]{\mbox{Xuhong Zhang}}
\author[1]{\mbox{Wenqi Zhang}}

\affiliation[1]{Zhejiang University}
\affiliation[2]{National University of Singapore}
\affiliation[3]{SAP}

\metadata[\faGlobe\ Project]{\url{https://zju-omniai.github.io/DTCA/}}
\metadata[\faGithub\ Code]{\url{https://github.com/ZJU-OmniAI/DTCA}}
\metadata[\faEnvelope\ Email]{\email{22651304@zju.edu.cn}}

\abstract{
Diffusion language models (DLMs) generate text through iterative denoising, enabling bidirectional context modeling and parallel token refinement. Recent studies have applied reinforcement learning (RL) to DLMs and improved their reasoning performance. However, the iterative denoising process makes credit assignment more challenging, as final outcome rewards provide limited information about how different tokens contribute to the final output. In this paper, we propose \textbf{Denoising Token Credit Assignment (DTCA)}, a framework for token-level credit assignment in reinforcement learning for diffusion language models. DTCA estimates token contributions by measuring the improvement in prediction confidence of remaining masked positions after token commitment and groups these signals according to part-of-speech (POS) categories. These contribution weights are used to select informative denoising timesteps and assign token-level learning signals during RL training. DTCA further calibrates these learning signals according to token-level entropy changes induced by policy updates. Experiments on LLaDA-8B-Instruct show that DTCA consistently improves mathematical reasoning performance across different generation budgets, with gains of up to 9.0 points on MATH500 and 5.3 points on GSM8K over the base model.
	
}

\hypersetup{bookmarksdepth=2,pdftitle={DTCA: Denoising Token Credit Assignment for Reinforcement Learning in Diffusion Language Models},pdfauthor={Hongze Chen, Miao Pan, Lizhou Liang, Qinfeng Li, Xinyin Ma, Yongliang Shen, Weishi Wang, Daniel Dahlmeier, Xuhong Zhang, Wenqi Zhang}}

\begin{document}
\thispagestyle{firstheader}
\maketitle
\pagestyle{empty}

\section{Introduction}

Diffusion language models (DLMs) have recently attracted increasing attention as an emerging paradigm for language modeling due to their ability to model bidirectional context and perform parallel token refinement~\citep{mercury,gemini_diffusion2025,song2025seed}, offering an alternative to autoregressive generation~\citep{llada,discrete_diffusion_lm,ye2025beyond,yu2025diffusionsurvey}. Recent studies have explored reinforcement learning for DLMs by leveraging the generation process, mainly constructing learning signals from denoising trajectories or intermediate states during training~\citep{rldf,tracerl,dispo,daca_grpo,dcolt}. However, DLMs generate sequences through multiple denoising steps, where tokens are gradually committed and each commitment can affect the refinement of
remaining masked positions~\citep{ou2024absorbing}. Therefore, RL optimization in DLMs requires finer-grained credit assignment that considers how different tokens affect subsequent refinement.
\begin{figure}[t]
	\centering
	\includegraphics[width=\linewidth]{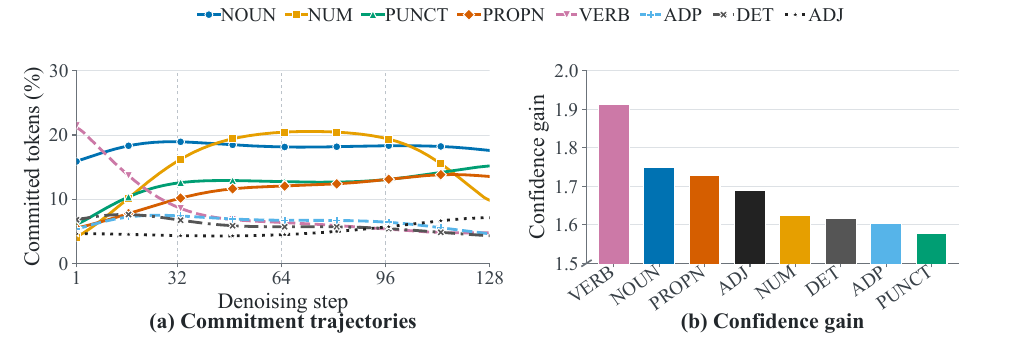}
    \caption{
    Preliminary analysis of POS-level differences in DLM denoising using rollouts
    generated from the GSM8K~\citep{cobbe2021gsm8k} training set.
    (a) Commitment timing of different POS categories across denoising steps.
    (b) Improvement in the prediction confidence of remaining masked
    positions after individual token commitment, averaged within each POS category.
    POS tags denote NOUN (noun), NUM (numeral), PUNCT (punctuation), PROPN (proper
    noun), VERB (verb), ADP (adposition), DET (determiner), and ADJ (adjective).
    }
	\label{fig:preliminary_observation}
\end{figure}

Achieving finer grained credit assignment requires estimating the contribution
of individual tokens during DLM denoising. Specifically, we measure token
contribution by how much a committed token improves the model's prediction
confidence on the remaining masked positions. However, directly estimating such
token-level confidence improvements during RL training is challenging because
multiple tokens are committed at the same denoising step and interact through
the shared denoising process, making their individual effects difficult to
isolate. Moreover, the estimated contribution can vary substantially across
different rollout samples. To obtain more reliable estimates, we categorize
tokens by part-of-speech (POS) categories and aggregate their refinement
effects within each category. POS categories provide a natural grouping criterion because they capture
different linguistic functions, with nouns and verbs typically carrying lexical
information while determiners and adpositions mainly providing grammatical
structure~\citep{de-marneffe-etal-2021-universal}.

The denoising behaviors of different POS categories are further analyzed using
rollouts generated from the GSM8K~\citep{cobbe2021gsm8k} dataset. As shown in Figure~\ref{fig:preliminary_observation}(a), different POS categories exhibit distinct commitment trajectories across denoising steps. NOUN maintains a relatively high commitment ratio, whereas VERB has a high ratio at the beginning that declines substantially over the denoising process. Other categories, such as ADP, DET, and ADJ, exhibit relatively lower commitment ratios with more stable variations. Figure~\ref{fig:preliminary_observation}(b) further shows the confidence gain associated with different POS categories. Clear differences emerge across categories: VERB achieves the largest confidence gain, followed by NOUN and PROPN, whereas categories such as DET and ADP show relatively smaller gains. These observations indicate that POS
categories capture different denoising behaviors and refinement effects, making
them a useful basis for grouping tokens and estimating contribution weights.

Based on these observations, we propose \textbf{Denoising Token Credit Assignment (DTCA)}, a token-level credit assignment framework for reinforcement learning in diffusion language models. DTCA estimates token contributions from denoising refinement signals and aggregates them into POS-level contribution weights to obtain more stable and interpretable learning signals. These contribution signals are then used to select informative denoising timesteps and derive token-level learning signals during RL optimization. DTCA further incorporates an entropy-based credit calibration module inspired by OPEFO~\citep{xu2026opefo} to calibrate the learning signals according to token-level entropy changes induced by policy updates. Experiments on LLaDA-8B-Instruct show that DTCA consistently improves mathematical reasoning performance across different generation budgets, with improvements of up to 9.0 percentage points on MATH500~\citep{lightman2024verify} and 5.3 percentage points on GSM8K~\citep{cobbe2021gsm8k} over the base model.

The main contributions are summarized as follows:
\begin{itemize}
    \item We analyze different denoising behaviors across POS categories in
    diffusion language models and use POS categories to estimate the effect of
    token commitments on subsequent refinement.
    \item We propose \textbf{DTCA}, a token-level credit assignment framework that aggregates denoising refinement signals by POS categories and uses the resulting contribution weights for denoising timestep selection and token credit assignment. DTCA further adapts entropy-based credit calibration from OPEFO~\citep{xu2026opefo} to the diffusion language model setting for improved RL optimization.
    \item  Extensive experiments demonstrate the effectiveness of DTCA on
    mathematical reasoning benchmarks and validate the benefits of its key
    components.
\end{itemize}

\section{Preliminaries}

\subsection{Diffusion Language Model Denoising}

Masked diffusion language models generate responses through iterative
denoising~\citep{llada,discrete_diffusion_lm}. Given a prompt $q$, let
$\tau=(x_T,x_{T-1},\ldots,x_0)$ denote a sampled trajectory, where $x_t$
is the partially masked response at denoising step $t$ and $x_0$ is the
completed response. Under the monotonic unmasking schedule used in this work,
each step reveals a subset of masked positions as predicted tokens; we call
this \emph{token commitment}. Committed tokens remain visible in subsequent
states and can influence predictions at the remaining masked positions. The
distribution $\pi_\theta(y\mid q)$ below is the response distribution induced
by this iterative sampling procedure, not an autoregressive token
factorization.

\subsection{Reinforcement Learning for Diffusion Language Models}

Given prompts $q\sim\mathcal{D}$ and sampled responses
$y\sim\pi_\theta(\cdot\mid q)$, reinforcement learning seeks to maximize
expected sequence-level reward:
\begin{equation}
    \mathcal{J}(\theta)
    =\mathbb{E}_{q\sim\mathcal{D},\,y\sim\pi_\theta(\cdot\mid q)}
    [R(q,y)].
\end{equation}
Group-relative policy optimization converts rewards from multiple responses
to the same prompt into response advantages~\citep{deepseekmath}. In masked
diffusion RL, these advantages can be applied to predictions at selected
denoising states~\citep{rldf}. Let $\mathcal{T}$ be the selected steps,
$\mathcal{B}_t$ their optimization target positions, and $v_i$ the target
token at position $i$. A schematic token-level surrogate is
\begin{equation}
    \mathcal{L}_{\mathrm{RL}}(\theta)
    =-\mathbb{E}\!\left[
        \frac{1}{N}\sum_{t\in\mathcal{T}}\sum_{i\in\mathcal{B}_t}
        A_{t,i}\log\pi_\theta(v_i\mid q,x_t)
      \right]
      +\beta_{\mathrm{KL}}\mathcal{L}_{\mathrm{KL}},
\end{equation}
where $N=\sum_{t\in\mathcal{T}}|\mathcal{B}_t|$ and
$\mathcal{L}_{\mathrm{KL}}$ denotes a penalty relative to a frozen reference
policy~\citep{deepseekmath,rldf}. The exact target set, credit
$A_{t,i}$, KL estimator, and update loss used by DTCA are specified below and
in Appendix~\ref{app:method_supplement}. A sequence-level reward alone does
not identify which commitments help subsequent denoising; this motivates
assigning credit at the denoising-state and target-token levels.

\section{Method}

\begin{figure}[t]
	\centering
	\includegraphics[width=\linewidth]{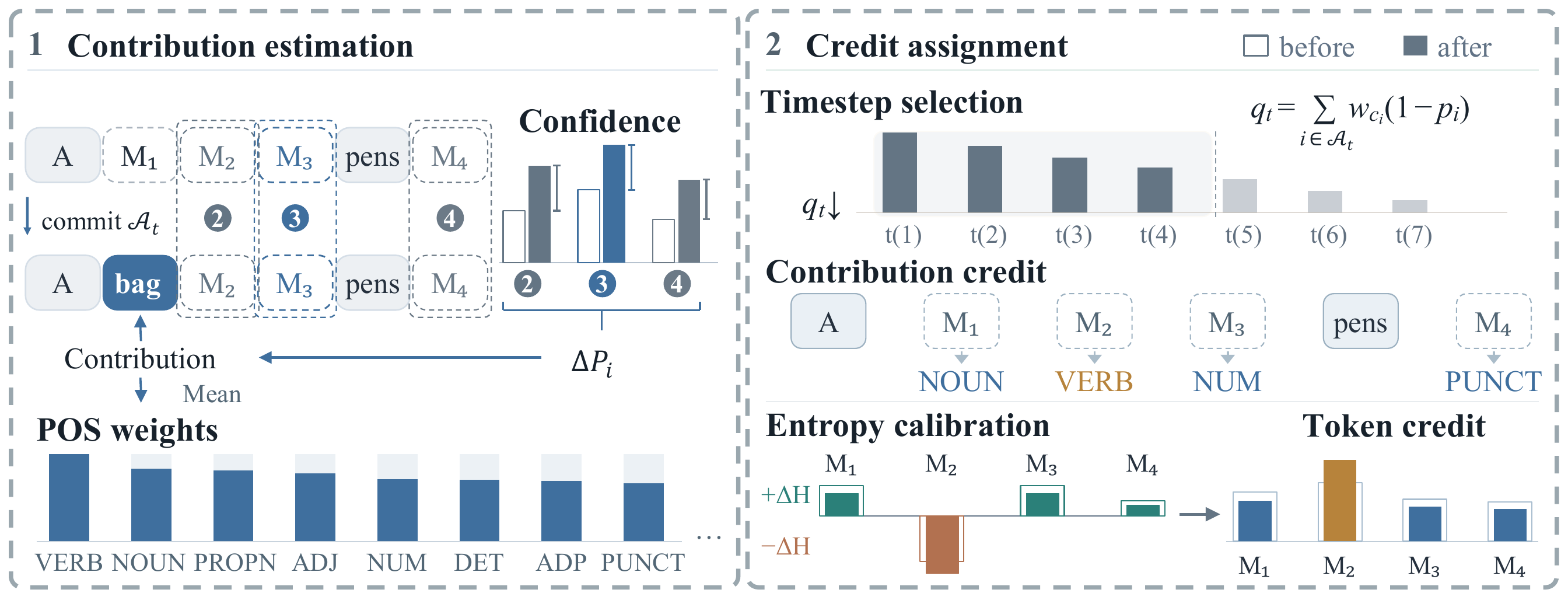}
	\caption{
		Overview of Denoising Token Credit Assignment (DTCA).
		DTCA estimates token contributions from denoising refinement dynamics and
		uses the resulting contribution weights for denoising state selection and
		token-level credit assignment. The obtained learning signals are further
		calibrated according to entropy changes induced by policy updates for
		reinforcement learning optimization.
	}
	\label{fig:dtca_overview}
\end{figure}

We propose \textbf{Denoising Token Credit Assignment (DTCA)}, a token credit
assignment framework for reinforcement learning in diffusion language models.
As illustrated in Figure~\ref{fig:dtca_overview}, DTCA estimates token
contributions from denoising refinement dynamics and uses the resulting
contribution weights for informative denoising state selection and token-level
credit assignment. DTCA further calibrates the token-level learning signals
according to entropy changes induced by policy updates.

\subsection{Token Contribution Estimation by POS Category}

As illustrated in Figure~\ref{fig:dtca_overview}, DTCA estimates token-level
refinement contributions from denoising dynamics and aggregates these estimates
into category-level contribution weights. Given a denoising trajectory
$\tau$, different token commitments may have different effects on subsequent
refinement. DTCA therefore estimates the refinement contribution of each
committed token by measuring the improvement in prediction confidence of the
remaining masked positions after the commitment.

Specifically, for each individual token commitment at denoising step $t$,
let $\mathcal{M}$ denote the response positions that remain masked both before
and after the commitment, excluding the committed position $i$. Given a fixed
prompt $q$, we define the aggregate prediction confidence of a denoising state
$x$ over $\mathcal{M}$ as the sum of the maximum prediction probabilities at those positions:
\begin{equation}
    F(x;\mathcal{M})
    =
    \sum_{j\in\mathcal{M}}
    \max_{v\in\mathcal{V}}p_{\theta,j}(v\mid q,x),
    \label{eq:confidence_definition}
\end{equation}
where $j$ indexes a masked position, $\mathcal{V}$ is the vocabulary, and
$p_{\theta,j}(v\mid q,x)$ is the probability assigned to token $v$ at position
$j$ conditioned on prompt $q$ and denoising state $x$. For a committed token
$v_i$, its refinement contribution estimate is computed as:
\begin{equation}
    \Delta P_i
    =
    \max\left(0,
    F(x_t^{(i)};\mathcal{M})-F(x_t;\mathcal{M})\right),
    \label{eq:token_contribution}
\end{equation}
where $x_t^{(i)}$ denotes the denoising state after committing token $v_i$
individually. Negative total changes are set to zero; the estimate is not
divided by the number of remaining masks. A larger $\Delta P_i$ indicates a stronger estimated refinement effect of the
committed token on subsequent denoising.

Obtaining reliable token-level estimates directly during RL training is
challenging because evaluating individual token effects requires re-evaluating
token commitments with additional forward passes, introducing substantial
computational overhead during policy optimization. Moreover, these estimates
can vary across different rollout samples. Therefore, DTCA constructs
contribution estimates from offline denoising rollouts before RL training and
groups them according to part-of-speech (POS) categories to obtain more stable
category-level contribution weights.

During offline estimation, DTCA uses a denoising budget equal to half of the
maximum generation length. At each denoising step, two tokens are committed,
and DTCA evaluates each token separately through additional forward passes
based on the confidence improvement of remaining masked positions. These estimates are aggregated at the POS level to reduce variance across rollouts.

For each committed token $v_i$, we assign its POS category as
$c_i=\phi(v_i)$. The contribution of each POS category is obtained by averaging
the refinement contribution estimates of tokens belonging to that category:
\begin{equation}
	m_c
	=
	\frac{1}{N_c}
	\sum_{i:c_i=c}
	\Delta P_i ,
	\label{eq:pos_contribution}
\end{equation}

where $N_c$ denotes the number of committed tokens in POS category $c$. The
resulting category-level contribution $m_c$ represents the average refinement
effect estimated for tokens in each category.

\paragraph{POS Contribution Weight Construction.}

The estimated POS contributions are normalized into bounded contribution
weights:
\begin{equation}
    w_c
    =
    0.8
    +
    0.4
    \frac{
        m_c-\min_{j\in\mathcal{C}}m_j
    }
    {
        \max_{j\in\mathcal{C}}m_j-\min_{j\in\mathcal{C}}m_j
    },
    \label{eq:pos_weight}
\end{equation}

where $\mathcal{C}$ denotes the set of POS categories. The weight
$w_c$ lies in the range $[0.8,1.2]$ and represents the scaling factor derived
from the estimated refinement contribution of category $c$. We constrain this
range to prevent excessive credit scaling while preserving the original RL
objective. These weights are used to adjust token-level credits during RL
optimization.

\subsection{Contribution-Based Timestep Selection}

The category contribution weights are also used to identify informative
denoising timesteps for RL training. Following RLDF~\citep{rldf}, DTCA
optimizes only a subset of denoising timesteps from each rollout. For a
denoising timestep $t$, let $\mathcal{A}_t$ denote the set of tokens committed
at this step. We compute the importance score of each timestep as:
\begin{equation}
	q_t
	=
	\sum_{i\in\mathcal{A}_t}
	w_{c_i}(1-p_i),
	\label{eq:timestep_score}
\end{equation}
where $p_i$ denotes the confidence of token $i$ when it is committed and
$w_{c_i}$ denotes the contribution weight of its POS category. The term
$1-p_i$ assigns higher scores to timesteps containing lower-confidence token
commitments, while $w_{c_i}$ incorporates the estimated refinement contribution
of the committed tokens.

DTCA ranks denoising timesteps according to $q_t$ and selects the smallest set
of timesteps whose cumulative score reaches a predefined fraction $\mu$ of the
total score. The number of selected timesteps therefore adapts to the
distribution of contribution signals in each rollout. Lower and upper bounds
are applied to the number of selected timesteps for stable and efficient
training.

\subsection{Contribution-Based Credit Assignment}

The category contribution weights are used to assign different credits to tokens
within each selected denoising timestep. For a selected timestep $t$, let
$\mathcal{B}_t$ denote the tokens used for optimization, including the committed
tokens at timestep $t$ and the retained masked positions selected by the
threshold-based filtering strategy in RLDF~\citep{rldf}. POS categories for all
tokens in $\mathcal{B}_t$ are obtained from the completed rollout, where
retained masked positions are associated with their final predicted tokens.
The credit weight of token $v_i$ is determined by the contribution weight of its
POS category:
\begin{equation}
	a_{t,i}
	=
	\frac{w_{c_i}}
	{\frac{1}{|\mathcal{B}_t|}
		\sum_{k\in\mathcal{B}_t}w_{c_k}},
	\label{eq:token_credit}
\end{equation}
where $c_i$ is the POS category of token $v_i$. The normalization keeps the average credit within each timestep equal to one, so that the overall scale of the timestep advantage remains unchanged. Given the original timestep advantage $A_t$, DTCA computes the token-level learning signal as
\begin{equation}
	A_{t,i}=a_{t,i}A_t .
	\label{eq:token_advantage}
\end{equation}
Tokens from categories with larger contribution weights receive stronger learning signals, while tokens from categories with smaller contribution receive weaker signals. In this way, DTCA transforms the original response-level feedback into more fine grained token credit during RL training.

\subsection{Entropy-Based Credit Calibration}

Although DTCA provides token-level learning signals according to denoising
contribution, different denoising states may induce different entropy changes
during policy updates. To further calibrate the token-level learning signals,
DTCA incorporates an entropy-based credit calibration module inspired by
OPEFO~\citep{xu2026opefo}. Unlike the original formulation for autoregressive
generation, we adapt the token-level entropy change estimation to selected
denoising states and their remaining target positions in diffusion language
models.

For each response $b$, selected denoising state $t$, and remaining target
position $i$, the entropy change induced by the policy update is estimated as:
\begin{equation}
    \Delta H_{t,i}^{(b)}
    =
    -\eta A_{t,i}^{(b)}
    (1-p_i^{(b)})^2
    (\log p_i^{(b)}+H_i^{(b)}),
    \label{eq:entropy_change}
\end{equation}

where $p_i^{(b)}$ denotes the probability assigned by the current policy to the
target token and $H_i^{(b)}$ denotes the entropy of the full prediction distribution
at this position. This formulation follows the first-order entropy change
estimation of OPEFO~\citep{xu2026opefo} while replacing autoregressive decoding steps with selected
denoising states and target positions.

Within each response, the entropy-increasing and entropy-decreasing updates
are separated according to the sign of $\Delta H_{t,i}^{(b)}$:
\begin{equation}
    S_b^{+}=\{(t,i)|\Delta H_{t,i}^{(b)}>0\},
    \qquad
    S_b^{-}=\{(t,i)|\Delta H_{t,i}^{(b)}<0\}.
    \label{eq:entropy_sets}
\end{equation}

Following OPEFO~\citep{xu2026opefo}, we compute a separate entropy flow
balancing coefficient for each response:
\begin{equation}
    \lambda_b
    =
    \operatorname{clip}
    \left(
    \frac{
    \sum_{(t,i)\in S_b^-}|\Delta H_{t,i}^{(b)}|
    -
    \sum_{(t,i)\in S_b^+}\Delta H_{t,i}^{(b)}
    }
    {
    \sum_{(t,i)\in S_b^-}|\Delta H_{t,i}^{(b)}|
    +
    \sum_{(t,i)\in S_b^+}\Delta H_{t,i}^{(b)}
    },
    -0.95,0.95
    \right).
    \label{eq:entropy_balance}
\end{equation}

The entropy balancing coefficient is then used to rescale token-level learning
signals to balance entropy-increasing and entropy-decreasing updates.
When $\lambda_b>0$, entropy-increasing updates are amplified and
entropy-decreasing updates are attenuated. When $\lambda_b<0$, the adjustment
is reversed. When $\lambda_b=0$, the credits remain unchanged:
\begin{equation}
    \hat{A}_{t,i}^{(b)}
    =
    \begin{cases}
    (1+\lambda_b)A_{t,i}^{(b)}, & \Delta H_{t,i}^{(b)}>0,\\
    (1-\lambda_b)A_{t,i}^{(b)}, & \Delta H_{t,i}^{(b)}\leq0 .
    \end{cases}
    \label{eq:entropy_rescale}
\end{equation}

\section{Experiments}

\subsection{Experimental Setup}
\label{sec:experimental_setup}

\textbf{Model and data.}
We initialize RL training from LLaDA-8B-Instruct~\citep{llada} without additional
SFT. Following RLDF~\citep{rldf}, we use the MATH~\citep{hendrycks2021math}
training split with the same data processing pipeline for RL optimization.
To construct the fixed POS contribution table, DTCA collects offline denoising
rollouts from the training data of MATH and GSM8K and builds a shared POS
contribution table without updating model parameters.

\textbf{Training.}
We train LLaDA-8B-Instruct with reinforcement learning without an intermediate
SFT stage. For each rollout, DTCA selects denoising states with the highest POS-weighted
uncertainty scores until covering 70\% of the total score mass, with the number of selected states constrained
between 2 and 32. We perform policy optimization using group-relative
advantages and regularize the policy against a frozen reference model with the
k3 KL estimator~\citep{Schulman_KLApprox}. Rollouts are generated using block-wise decoding with KV cache, following Fast-dLLM~\citep{fastdllm}. Full implementation details and
hyperparameters are provided in Appendix~\ref{app:experimental_setup}.

\textbf{Evaluation and baselines.}
We evaluate mathematical reasoning performance on MATH500~\citep{lightman2024verify}
and GSM8K~\citep{cobbe2021gsm8k} using final-answer accuracy. Following prior
work, we evaluate all models under generation budgets of 128, 256, and 512,
with denoising steps matched to the corresponding maximum generation lengths.
We compare DTCA with LLaDA-8B-Instruct, Uniform SFT, LLaDA-1.5, and
representative reinforcement learning methods for diffusion language models,
including d1~\citep{zhao2025d1}, ESPO~\citep{espo}, TraceRL~\citep{tracerl},
Coupled-GRPO~\citep{diffucoder}, and RLDF~\citep{rldf}. Additional details on
training configurations and baseline evaluation settings are provided in
Appendix~\ref{app:training_configuration} and Appendix~\ref{app:evaluation_settings}.

\subsection{Main Results}

Table~\ref{tab:main_results} compares DTCA with supervised and reinforcement
learning baselines on MATH500 and GSM8K. DTCA consistently improves over
LLaDA-8B-Instruct across both benchmarks and all evaluated generation budgets.
On MATH500, DTCA improves the base model by 9.0, 6.8, and 4.8 percentage points
at budgets of 128, 256, and 512, respectively. On GSM8K, the corresponding
improvements are 5.3, 2.1, and 2.5 percentage points. These results show that
DTCA provides consistent gains under different denoising budgets rather than
relying on a specific inference setting. Compared with supervised optimization
and existing diffusion language model reinforcement learning methods, DTCA
achieves competitive performance across the reported settings, demonstrating
the effectiveness of the proposed token-level credit assignment framework. 

The largest improvements on both benchmarks occur at the smallest evaluated
generation budget of 128, while gains remain positive at budgets of 256 and
512. Although the relative improvements become smaller at larger generation
budgets, DTCA consistently improves over the base model, highlighting its
stronger gains under limited denoising computation.
\begin{table}[t]
	\centering
	\small
	\caption{
		Mathematical reasoning accuracy (\%) on MATH500 and GSM8K.
		The last rows report DTCA, its ablation variants, and the improvement over the
		base model. Boldface indicates the best result in each column, and $\Delta$
		denotes the improvement over LLaDA-8B-Instruct.
	}
	\label{tab:main_results}
	\setlength{\tabcolsep}{4pt}
	\begin{tabular*}{\linewidth}{@{\extracolsep{\fill}}lcccccc@{}}
		\toprule
		& \multicolumn{3}{c}{MATH500}
		& \multicolumn{3}{c}{GSM8K} \\
		\cmidrule(lr){2-4}\cmidrule(lr){5-7}
		Method & 128 & 256 & 512 & 128 & 256 & 512 \\
		\midrule
		LLaDA-8B-Instruct~\citep{llada}
		& 30.8 & 36.4 & 39.2
		& 75.3 & 83.2 & 82.9 \\
		
		LLaDA-1.5~\citep{llada15}
		& 32.6 & 39.4 & 39.8
		& 76.2 & 81.6 & 83.0 \\
		
		Uniform SFT
		& 24.6 & 30.0 & 32.6
		& 74.8 & 80.7 & 78.6 \\
		
		\midrule
		
		d1~\citep{zhao2025d1}
		& 33.6 & 37.6 & 37.2
		& 75.8 & 82.4 & 81.8 \\
		
		ESPO~\citep{espo}
		& 37.6 & 38.0 & 40.2
		& 78.1 & 83.2 & 84.0 \\
		
		TraceRL~\citep{tracerl}
		& 32.6 & 34.2 & 34.0
		& 76.6 & 80.4 & 81.8 \\
		
		Coupled-GRPO~\citep{diffucoder}
		& 35.2 & 40.2 & 40.2
		& 78.0 & 83.6 & 83.2 \\
		
		RLDF~\citep{rldf}
		& 36.2 & 41.2 & 42.6
		& 77.2 & 83.4 & 82.7 \\
		
		\midrule
		
		DTCA w/o Entropy Calibration
		& 37.4 & 41.8 & 39.8
		& 78.3 & 83.9 & 81.9 \\
		
		DTCA w/o Contribution Credit
		& 36.2 & 39.4 & 41.2
		& 77.1 & 82.9 & 82.3 \\
		
		DTCA (ours)
		& \textbf{39.8} & \textbf{43.2} & \textbf{44.0}
		& \textbf{80.6} & \textbf{85.3} & \textbf{85.4} \\
		
		$\Delta$ vs.\ Base
		& +9.0 & +6.8 & +4.8
		& +5.3 & +2.1 & +2.5 \\
		
		\bottomrule
	\end{tabular*}
\end{table}

\begin{figure}[t]
    \centering
    \includegraphics[width=\linewidth]{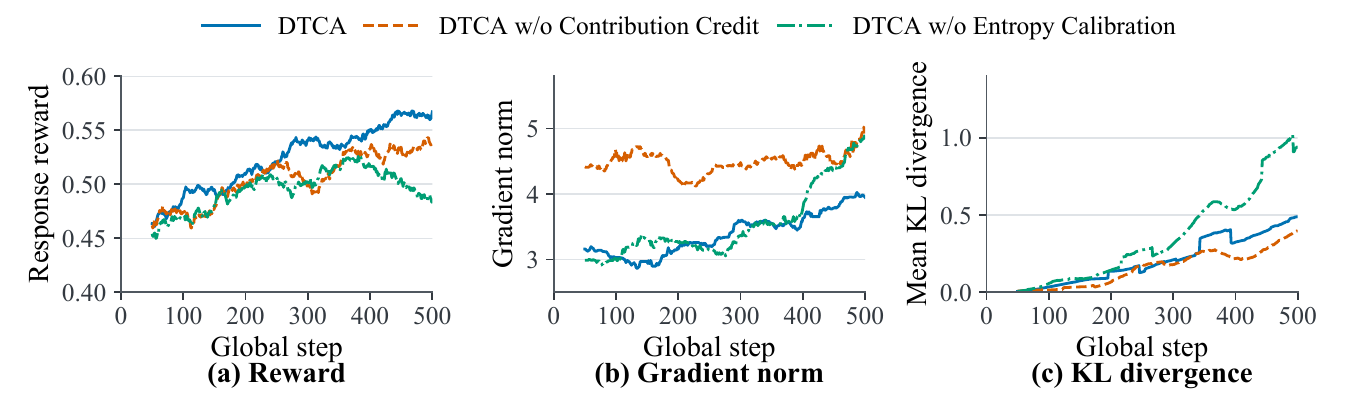}
    \caption{
    Optimization dynamics of DTCA variants during RL training.
    (a) Response reward, (b) gradient norm, and (c) mean KL divergence.
    DTCA is compared with variants without Contribution Credit and Entropy Calibration.
    }
    \label{fig:dtca_ablation_dynamics}
\end{figure}

\subsection{Ablation Study}

Table~\ref{tab:main_results} reports ablation results for different DTCA
components. \textit{DTCA w/o Contribution Credit} replaces the
contribution-based token credits with uniform credits while keeping the same
timestep selection strategy. \textit{DTCA w/o Entropy Calibration} removes the
entropy-based credit calibration module while preserving contribution-based
credit assignment and timestep selection.

Removing either component decreases performance compared with the full DTCA
model across both benchmarks and different generation budgets. At the 256
generation budget, removing Contribution Credit leads to decreases of 3.8 and
2.4 percentage points on MATH500 and GSM8K, respectively, while removing
Entropy Calibration results in smaller drops of 1.4 and 1.4 points. At the 512
generation budget, removing Entropy Calibration causes larger performance
decreases, with drops of 4.2 and 3.5 points on MATH500 and GSM8K. These results
show that both components contribute to the overall performance of DTCA.

\subsection{Analysis of Optimization Dynamics}

To understand how different DTCA components influence RL optimization, we
analyze training dynamics in terms of reward, gradient norm, and KL divergence.
Figure~\ref{fig:dtca_ablation_dynamics} compares DTCA with variants without
Contribution Credit and Entropy Calibration during training.

Removing Contribution Credit results in similar reward trajectories to DTCA,
but leads to substantially different update behaviors. In particular, the
variant with uniform credit assignment consistently exhibits larger gradient
norms throughout training while achieving comparable reward improvements. This
suggests that contribution-based credit assignment mainly affects the quality
of token-level learning signals, allowing the model to optimize with different
update dynamics rather than simply changing the final reward trend. By
considering token-level contribution differences, DTCA provides a more
structured learning signal than uniform credit assignment.

Removing Entropy Calibration mainly affects the stability of later-stage policy
updates. Although its reward trajectory initially follows a similar trend, the
variant without entropy calibration shows a faster increase in KL divergence
and lower reward in later training stages. This indicates that entropy
calibration helps control policy changes by adjusting learning signals
according to entropy changes, leading to more stable optimization during RL
training.

\begin{figure}[t]
    \centering
    \includegraphics[width=\linewidth]{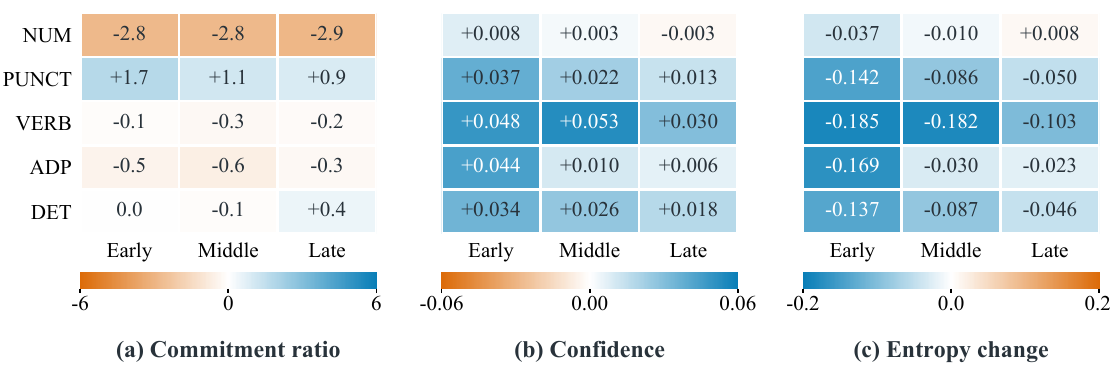}
\caption{
	Analysis of commitment dynamics on MATH500.
	DTCA is compared with the base model across three denoising stages. Each cell
	shows the difference in (a) token commitment ratio, (b) confidence at
	commitment, and (c) entropy at commitment.
}
    \label{fig:token_commitment_analysis}
\end{figure}

\subsection{Analysis of Token Commitment Dynamics}

To understand whether contribution-based credit assignment changes the denoising
behavior of diffusion language models, we analyze token commitment patterns after
RL training. Token commitment reflects when tokens become fixed during
denoising, while the commitment distribution and prediction states provide
insights into the learned denoising behavior. Figure~\ref{fig:token_commitment_analysis}
compares DTCA with the base model on MATH500 across early, middle, and late
denoising stages. Each cell reports the difference between DTCA and the base
model in token commitment ratio, prediction confidence at commitment, and target
entropy at commitment.

DTCA changes the commitment distribution across token categories rather than
uniformly shifting all commitments. For example, DTCA consistently reduces the
commitment ratio of NUM tokens while increasing the commitment ratio of PUNCT
tokens across different denoising stages. These changes suggest that DTCA
learns different commitment patterns across token categories instead of simply
accelerating or delaying all token predictions. Notably, changes in commitment
frequency do not always correspond to changes in commitment states. For
example, VERB shows limited changes in commitment ratio but substantial
increases in confidence and reductions in entropy at commitment. This indicates
that DTCA influences not only which tokens are committed, but also the model
states when commitments occur. The observed changes are consistent with
leveraging denoising refinement signals to improve token-level credit assignment
during reinforcement learning.

\subsection{Analysis of Entropy-Based Credit Calibration}

\begin{figure}[t]
    \centering
    \includegraphics[width=\linewidth]{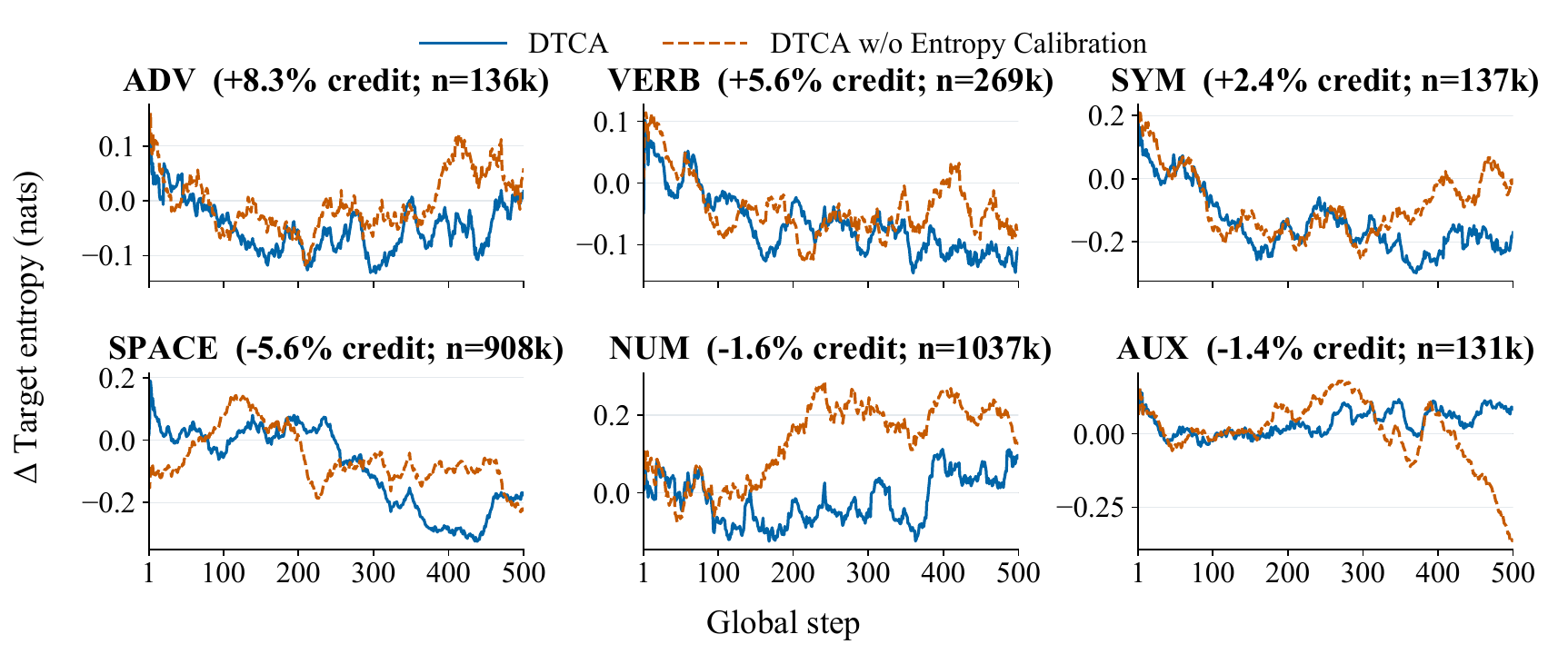}
    \caption{
    Analysis of entropy-based credit calibration.
    Each panel shows the target entropy change of a POS category during RL
    training. The blue solid and orange dashed curves represent DTCA and DTCA
    without entropy calibration, respectively. The vertical axis shows the
    difference between the mean predictive entropy over the most recent 25
    updates and the mean over the first 100 updates of the same run. Percentages
    in parentheses show the relative change in assigned credit after entropy
    calibration, and $n$ is the number of tokens counted.
    }
    \label{fig:entropy_calibration_analysis}
\end{figure}

We study how entropy calibration changes token credits and prediction entropy
across POS categories. We compare DTCA with a variant that keeps the same
contribution-based credit assignment and timestep selection but removes
entropy calibration. Figure~\ref{fig:entropy_calibration_analysis} shows the
entropy changes during training. Positive values mean increased entropy, and
negative values mean decreased entropy. Values in parentheses show the
relative change in assigned credit after calibration. We compare these credit
changes with the entropy curves for each category.

In Figure~\ref{fig:entropy_calibration_analysis}, the two variants show similar
entropy changes early in training and gradually diverge later. For most of
the shown POS categories, DTCA has a lower curve late in training. Each curve
shows changes relative to that run's early training average. For example,
without calibration, the NUM curve rises above zero after about 200 updates
and stays positive for most of the remaining steps. DTCA stays near or below
zero for most of this period. For SYM, the curve without calibration returns
toward zero late in training, while DTCA remains negative. AUX shows the
opposite pattern. Its curve without calibration drops sharply below zero near
the end, while the DTCA curve stays close to zero. The differences between
the two variants therefore vary across POS categories, rather than following
the same direction for all categories.

Credit changes and entropy changes do not always follow the same direction.
For example, SYM receives more credit but stays below its early entropy
average late in training. AUX receives less credit but ends above its early
average. Credit changes the strength of the existing policy update. Whether
entropy rises or falls also depends on the advantage signal, the current
prediction distribution, and the target token. Compared with
OPEFO~\citep{xu2026opefo}, DTCA applies entropy flow balancing after assigning
POS-based credits. Within each response, updates in the same entropy-change
set share a scaling factor. This factor and the final normalization preserve
the credit ratios within that set. Thus, entropy flow balancing still leaves
different credits for tokens with different POS contribution weights. The
two runs also see different denoising states and different mixes of target
tokens as training progresses. The curves therefore show overall training
trends, rather than measuring the same tokens immediately before and after a
single policy update.

\section{Conclusion}

We presented \textbf{Denoising Token Credit Assignment (DTCA)}, a framework for
fine-grained token-level credit assignment in reinforcement learning for
diffusion language models. DTCA estimates token contributions from the
confidence improvement of remaining masked positions after token commitment,
aggregates these signals by POS categories, and uses the resulting contribution
weights for denoising timestep selection and token-level credit assignment.
DTCA further calibrates contribution credits according to entropy changes
induced by policy updates, balancing entropy-increasing and entropy-decreasing
updates. On LLaDA-8B-Instruct, DTCA improves the reproduced base model by 6.8
percentage points on MATH500 and 2.1 points on GSM8K under the 256-step
evaluation setting. Training dynamics and component ablations show that
contribution-based credit assignment improves token-level learning signals,
while entropy-based calibration helps maintain more stable policy updates.
These results demonstrate that denoising dynamics provide useful information
for fine-grained credit assignment in reinforcement learning for diffusion
language models.

\FloatBarrier
\bibliographystyle{assets/acl_natbib}
\bibliography{paper}

\clearpage
\beginappendix

\raggedbottom
\setlength{\floatsep}{10pt plus 2pt minus 2pt}
\setlength{\textfloatsep}{10pt plus 2pt minus 2pt}
\setlength{\intextsep}{10pt plus 2pt minus 2pt}
\captionsetup{skip=4pt}

\section{Related Work}

\subsection{Diffusion Language Models}

Building on the denoising paradigm of diffusion models~\citep{ho2020ddpm}, diffusion language models (DLMs) generate text through iterative denoising~\citep{hoogeboom2021argmax,dieleman2022continuous,gulrajani2023likelihood} rather than autoregressive next-token prediction~\citep{lou2024sedd,li2022diffusionlm,diffusion_of_thought}.
Early discrete diffusion language models explore masked generation objectives and iterative refinement for language modeling~\citep{austin2021structured,discrete_diffusion_lm,shi2024simplified}.
LLaDA scales this paradigm to large language models with a masked diffusion objective and demonstrates competitive language generation performance~\citep{llada}.
Subsequent works further study improved training objectives, generation strategies~\citep{wu2025fastv2,xie2025dreamcoder}, scaling behaviors~\citep{dream7b,diffucoder,fastdllm}, and data efficiency under limited-data and multi-epoch training settings~\citep{prabhudesai2025dataconstrained,ni2025superdata,gao2025superdata}.
When applying reinforcement learning to these models, their iterative generation process requires particular consideration.
Unlike autoregressive models, DLMs maintain multiple intermediate denoising states, where tokens are progressively committed and remaining masked positions are refined.
This generation process introduces new challenges and opportunities for reinforcement learning in diffusion models.

\subsection{Reinforcement Learning for Diffusion Language Models}

Reinforcement learning for diffusion language models requires optimization
methods that account for the iterative denoising process~\citep{mask_aware_policy_gradients,mdpo}. Existing approaches
mainly adapt RL objectives~\citep{schulman2017ppo,deepseekmath,sepo,ni2026justgrpo} or exploit intermediate denoising information~\citep{denoising_process_rewards} to
construct effective learning signals~\citep{wd1,gdpo,llada15}. d1 introduces diffu-GRPO ~\citep{zhao2025d1}, Coupled-GRPO reduces likelihood
estimation variance through complementary masks~\citep{diffucoder}, and ESPO
uses evidence lower bound as a proxy for sequence likelihood during RL
optimization~\citep{espo}. Other methods leverage denoising trajectories~\citep{cj_grpo} and
intermediate states to improve RL training, including TraceRL, RLDF, and
DiSPO~\citep{tracerl,rldf,dispo}. These methods demonstrate how diffusion
generation characteristics can be incorporated into RL optimization~\citep{diffpo}. However,
they mainly focus on optimization objectives~\citep{stabledrl,dipod}, likelihood estimation~\citep{spg}, or
denoising trajectories, while how to assign fine-grained credit to individual
tokens during the denoising process remains underexplored.

\subsection{Fine-grained Credit Assignment}

Sequence-level rewards in language model reinforcement learning~\citep{instructgpt,deepseekr1} provide limited
information about which tokens or reasoning steps contribute to the final
outcome, motivating recent studies on fine-grained credit assignment~\citep{high_entropy_tokens,cheng2026exploration}. Existing
approaches improve reward granularity through process supervision~\citep{uesato2022process}, step-level
feedback, and token-level reward redistribution~\citep{gtpo} or importance estimation
~\citep{lightman2024verify,math_shepherd,red,prime,vineppo,gcpo,delta_credit}.
However, these methods are mainly developed for autoregressive language models,
where tokens are generated sequentially. For diffusion language models,
DACA-GRPO provides an initial exploration of token-level credit assignment by
weighting committed tokens according to denoising progress and changes in the
predicted probabilities of remaining masked positions~\citep{daca_grpo}.
Different from DACA-GRPO, DTCA estimates token contributions from confidence
improvements after token commitment, aggregates these signals according to POS
categories, and uses the resulting contribution structure for denoising
timestep selection and token-level credit assignment.

\section{Method Supplement}
\label{app:method_supplement}

\paragraph{Notation and boundary cases.}
In Eq.~\ref{eq:timestep_score}, $\mathcal{A}_t$ contains the positions
committed at denoising step $t$ and is used for timestep selection. The
optimization targets $\mathcal{B}_t$ may differ and include both committed
positions and retained masked positions selected by threshold-based filtering.
POS categories for all optimization targets are obtained from the corresponding
tokens in the completed rollout. In Eq.~\ref{eq:entropy_change}, $\eta$ denotes
the policy update step size used in the first-order entropy change approximation,
and $S_b^+$ and $S_b^-$ collect targets with positive and negative estimated
entropy changes in response $b$.

All 18 POS categories have entries in the contribution table. A POS tag outside
this inventory, or a model token that cannot be aligned to a POS span, is
assigned to the \texttt{X} category and uses its computed weight
$w_{\mathrm{X}}$. If all category means coincide, we set $w_c=1$ for every
category. If the total magnitude of estimated entropy change is zero, we set
$\lambda_b=0$. A zero response advantage yields zero token credits without
applying Eq.~\ref{eq:entropy_normalization}. We exclude empty commitment sets
from offline statistics and states with no optimization targets from training.

\begin{algorithm}[H]
\caption{One RL Update with Denoising Token Credit Assignment (DTCA)}
\label{alg:dtca}
\small
\begin{algorithmic}[1]
\State \textbf{Input:} Policy $\pi_\theta$, reference $\pi_{\mathrm{ref}}$, prompts $\mathcal{X}$, optimizer $O$
\Statex \hspace{\algorithmicindent} Fixed offline POS weights $\{w_c\}$, threshold $\mu$
\State $\tau\leftarrow\pi_\theta(\mathcal{X})$ \Comment{Generate denoising rollouts}
\State Compute rewards, advantages $\{A^{(b)}\}$, and token POS categories
\For{each response $b$ in $\tau$}
    \For{each denoising timestep $t$ in response $b$}
        \State $q_t^{(b)}\leftarrow\sum_{i\in\mathcal{A}_t^{(b)}}w_{c_i}(1-p_i)$
    \EndFor
    \State Select the highest-scoring steps covering mass $\mu$; apply count bounds to obtain $\mathcal{T}^{(b)}$
    \For{each selected timestep $t\in\mathcal{T}^{(b)}$}
        \State Reconstruct $x_t^{(b)}$ and form the retained targets $\mathcal{B}_t^{(b)}$
        \State $a_{t,i}^{(b)}\leftarrow w_{c_i}/\mathrm{Mean}_{k\in\mathcal{B}_t^{(b)}}(w_{c_k})$ for $i\in\mathcal{B}_t^{(b)}$
        \State $A_{t,i}^{(b)}\leftarrow a_{t,i}^{(b)}A^{(b)}$
        \State Estimate $\Delta H_{t,i}^{(b)}$ from $A_{t,i}^{(b)}$, current target probabilities, and entropies
    \EndFor
\EndFor
\For{each response $b$ in $\tau$}
    \State Form $S_b^+$ and $S_b^-$; compute $\lambda_b$ and clip it to $[-0.95,0.95]$
    \State Scale $A_{t,i}^{(b)}$ by $1+\lambda_b$ if $\Delta H_{t,i}^{(b)}>0$, otherwise by $1-\lambda_b$, to obtain $\hat{A}_{t,i}^{(b)}$
    \State Normalize $\hat{A}_{t,i}^{(b)}$ to obtain $\widetilde{A}_{t,i}^{(b)}$ \Comment{Preserve total response credit}
\EndFor
\State $\theta\leftarrow O(\theta,\nabla_\theta\mathcal{L}_{\mathrm{DTCA}})$
\end{algorithmic}
\end{algorithm}

\paragraph{Training objective.}
Let $A^{(b)}$ be response $b$'s group-normalized advantage, used as $A_t$ in
Eq.~\ref{eq:token_advantage} at each selected step
$t\in\mathcal{T}^{(b)}$. The targets $\mathcal{B}_t^{(b)}$ include masked
positions with predicted token probability at least $0.2$ and positions
committed at step $t$. Since the completed rollout is available after
generation, POS categories can be assigned to all targets in
$\mathcal{B}_t^{(b)}$. Eqs.~\ref{eq:token_credit}--\ref{eq:entropy_rescale}
yield $\hat{A}_{t,i}^{(b)}$ for each target. We restore the response's total
credit:
\begin{equation}
	\widetilde{A}_{t,i}^{(b)}
	=
	\hat{A}_{t,i}^{(b)}
	\frac{
		\sum_{t'\in\mathcal{T}^{(b)}}\sum_{j\in\mathcal{B}_{t'}^{(b)}}A_{t',j}^{(b)}
	}
	{
		\sum_{t'\in\mathcal{T}^{(b)}}\sum_{j\in\mathcal{B}_{t'}^{(b)}}\hat{A}_{t',j}^{(b)}
	}.
	\label{eq:entropy_normalization}
\end{equation}

This response-wide factor restores the overall credit scale after entropy
calibration, while clipping $\lambda_b$ may leave a residual entropy
imbalance. The update loss is
\begin{equation}
	\mathcal{L}_{\mathrm{DTCA}}(\theta)
	=\frac{1}{N}\sum_b\sum_{t\in\mathcal{T}^{(b)}}
	\sum_{i\in\mathcal{B}_t^{(b)}}
	\left[-\widetilde{A}_{t,i}^{(b)}
	\log\pi_\theta(v_i^{(b)}\mid q^{(b)},x_t^{(b)})
	+\beta_{\mathrm{KL}}\left(e^{d_{t,i}^{(b)}}-1-d_{t,i}^{(b)}\right)\right],
	\label{eq:dtca_training_objective}
\end{equation}
where $N=\sum_b\sum_{t\in\mathcal{T}^{(b)}}|\mathcal{B}_t^{(b)}|$ and
$d_{t,i}^{(b)}=\log\pi_{\mathrm{ref}}(v_i^{(b)}\mid q^{(b)},x_t^{(b)})
-\log\pi_\theta(v_i^{(b)}\mid q^{(b)},x_t^{(b)})$.
The reference policy is frozen. With one policy iteration, PPO ratio clipping
is inactive. Entropy calibration modifies $\widetilde{A}_{t,i}^{(b)}$ rather
than adding an entropy loss.

\section{Experimental Setup and Implementation Details}
\label{app:experimental_setup}

\subsection{Base Models and Datasets}
\label{app:base_models_datasets}

The policy and frozen reference model both start from LLaDA-8B-Instruct,
without an intermediate SFT stage. RL training uses the MATH training set
processed for RLDF. Offline POS contribution
estimates are constructed from denoising rollouts over the MATH and GSM8K
training splits without parameter updates to build a shared fixed POS
contribution table. We evaluate on MATH500 and the GSM8K test split; neither
evaluation split is used for RL training or POS table construction.

\subsection{Training Configuration}
\label{app:training_configuration}

We reproduce the compared methods under a unified experimental setup wherever
applicable. RL methods start from LLaDA-8B-Instruct and use the same
RLDF-processed MATH training split and mathematical correctness reward. All
methods are evaluated using the common protocol in
Appendix~\ref{app:evaluation_settings}. Each method retains its own objective
and method-specific settings: we follow its original paper first, and use its
released open-source code for details not specified in the paper.

For DTCA and the locally reproduced RLDF baseline, we use full-parameter
bfloat16 training with AdamW, a constant learning rate of $10^{-6}$, and zero
weight decay. Each prompt produces four responses, whose rewards are
standardized within the group to form response advantages. Both methods use the same low-probability threshold of 0.2 for training-target
filtering, token-level loss normalization, and a frozen-reference $k_3$ KL
penalty with coefficient $0.01$.

Their training rollouts use a maximum response length of 256 tokens and 256
denoising steps. Decoding uses a block size of 32 and a dynamic confidence
threshold of $0.9$; when no position crosses the threshold, the
highest-confidence position is committed. This decoding threshold is distinct
from the $0.2$ training-target filter. KV caching is used during rollout
generation.

\subsection{Implementation Details of DTCA}
\label{app:dtca_implementation}

We tag completions with spaCy's \texttt{en\_core\_web\_sm} POS model and align
its word spans with model token positions. The 18 native POS categories share
one contribution table. Before RL training, we map the offline category means to
$[0.8,1.2]$ using Eq.~\ref{eq:pos_weight}. These weights are used for both
timestep selection and token credit assignment and remain fixed throughout RL training.

DTCA changes the timestep-selection and credit-assignment rules relative to
the locally reproduced RLDF baseline. For state selection, we rank observed
denoising steps by $q_t$ and cover at least $70\%$ of the total priority mass,
selecting between 2 and 32 steps. At each selected step, training targets are
the remaining masked positions retained by the shared low-probability filter,
plus the positions committed at that step even if they fall below the filter
threshold. POS credits are normalized to preserve the mean response advantage
within each selected step. Entropy-flow calibration limits the absolute
balancing coefficient to $0.95$ and restores each response's total credit
after rescaling.

\subsection{Evaluation Settings}
\label{app:evaluation_settings}

We evaluate each checkpoint at maximum response lengths of 128, 256, and
512 tokens, using the same number of denoising steps as the response length.
For every setting, we use block size 32, dynamic confidence threshold $0.9$,
temperature $0$, and one generated response per question. The same math
prompt template and decoding settings are used for DTCA and all locally
reproduced baselines.

We evaluate only whether the final answer is correct, without scoring the
intermediate reasoning steps. Accuracy is the fraction of questions with
correct final answers on the complete test
split. All baseline scores in Table~\ref{tab:main_results} are obtained from
our local reproductions under this evaluation protocol.

\section{Additional Experiments and Analysis}
\label{app:additional_experiments}

\subsection{Additional Ablation Studies}
\label{app:additional_ablation}

\paragraph{Contribution-based timestep selection.}
Table~\ref{tab:appendix_timestep_selection} compares selection rules on a
shared dataset using our fixed POS weights in $[0.8,1.2]$. DTCA uses
$\mu\in\{0.40,0.70\}$ and selects 2--32 states. RLDF samples without
replacement from a temperature-1 softmax of state mean token surprisal,
using 16 or 32 states or DTCA's per-response count. RLDF top-$k$ takes the
$k$ highest-surprisal states, or all available states if fewer than $k$.
Coverage is the fraction of committed tokens, confidence $\sum_i p_i$,
low-confidence mass $\sum_i(1-p_i)$, or entropy $\sum_i H_i$ in selected
states. Ratios are averaged over responses. Sampled RLDF uses 2,000
independent selections.

\begin{table}[H]
    \centering
    \small
    \caption{Coverage of selected denoising states. State ratio is the
    selected fraction per response. RLDF (same count) matches DTCA's count
    at each $\mu$. All entries are mean percentages.}
    \label{tab:appendix_timestep_selection}
    \setlength{\tabcolsep}{4pt}
    \begin{tabular*}{\linewidth}{@{\extracolsep{\fill}}lccccc@{}}
        \toprule
        Method & State ratio & Token cov. & Conf. cov. & Low conf. cov. & Entropy cov. \\
        \midrule
        DTCA ($\mu=0.40$) & 16.82 & 19.99 & 17.72 & 40.59 & 37.81 \\
        RLDF $k=16$ & 23.31 & 21.33 & 20.79 & 26.52 & 25.56 \\
        RLDF (same count, $\mu=0.40$) & 16.82 & 15.01 & 14.34 & 19.93 & 19.22 \\
        RLDF top-16 & 23.31 & 10.01 & 7.33 & 42.46 & 35.88 \\
        \midrule
        DTCA ($\mu=0.70$) & 31.81 & 36.02 & 33.24 & 62.81 & 59.31 \\
        RLDF $k=32$ & 45.42 & 42.61 & 41.73 & 49.91 & 48.73 \\
        RLDF (same count, $\mu=0.70$) & 31.81 & 28.94 & 28.06 & 36.46 & 35.21 \\
        RLDF top-32 & 45.42 & 28.75 & 24.99 & 66.39 & 60.66 \\
        \bottomrule
    \end{tabular*}
\end{table}

At a matched state ratio of 16.82\%, DTCA ($\mu=0.40$) increases
low-confidence coverage from 19.93\% to 40.59\% and entropy coverage from
19.22\% to 37.81\% relative to RLDF, while retaining broader token coverage.
RLDF top-16 achieves slightly higher low-confidence coverage (42.46\%) but
covers fewer tokens (10.01\% versus 19.99\%).

At $\mu=0.70$ and a matched state ratio of 31.81\%, DTCA increases
low-confidence coverage from 36.46\% to 62.81\% and entropy coverage from
35.21\% to 59.31\%. It also achieves higher coverage than RLDF $k=32$ while
selecting fewer states. RLDF top-32 obtains comparable uncertainty coverage
but selects more states and covers fewer tokens. These results show that DTCA
selects denoising states with different uncertainty distributions, while the
impact of state selection on final performance is evaluated through the main
results and ablations.

\Needspace{9\baselineskip}
\subsection{Analysis of Denoising Dynamics}
\label{app:denoising_dynamics_analysis}

\begin{figure}[H]
    \centering
    \includegraphics[width=0.90\linewidth]{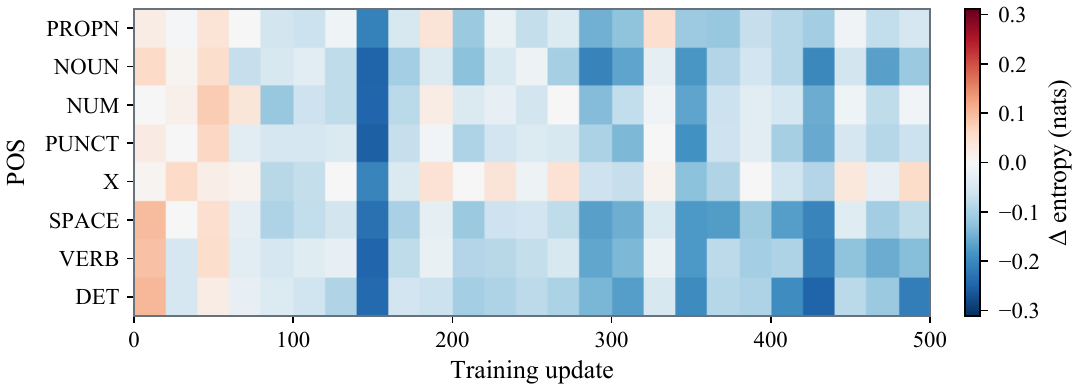}
    \caption{Current-commit token entropy changes in DTCA. Each 20-update
    bin subtracts the POS-specific mean over updates 1--100. The color scale
    matches the upper panel of Figure~\ref{fig:appendix_entropy_training}.}
    \label{fig:appendix_pos_policy_entropy_dtca}
\end{figure}

\paragraph{Policy entropy changes across POS categories.}
Figure~\ref{fig:appendix_pos_policy_entropy_dtca} and the upper panel of
Figure~\ref{fig:appendix_entropy_training} compare current-commit
token entropy by POS. Over updates 401--500, NUM entropy is 0.063 nats below
its early baseline in DTCA but 0.208 nats above it without calibration.
Late positive shifts in the uncalibrated run also appear for PROPN, X, and
SPACE, rather than being confined to NUM.

\Needspace{7\baselineskip}
\subsection{Additional Results}
\label{app:additional_results}

\begin{figure}[H]
    \centering
    \includegraphics[width=0.90\linewidth]{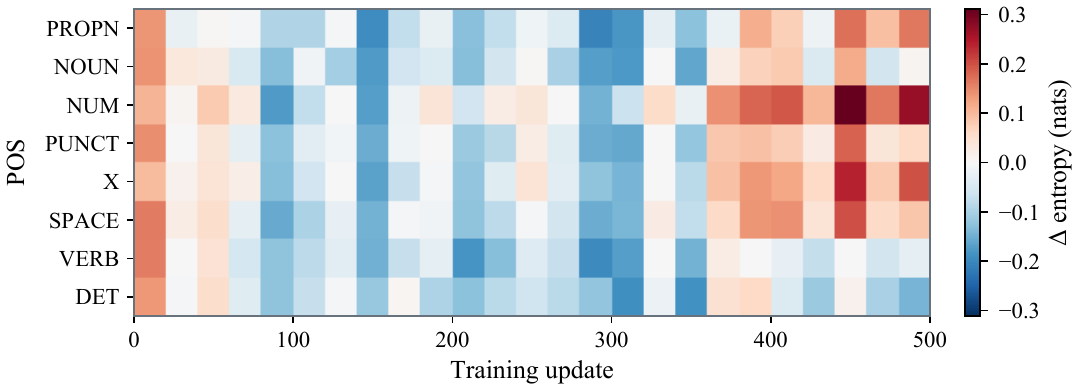}
    \par\vspace{5pt}
    \includegraphics[width=\linewidth]{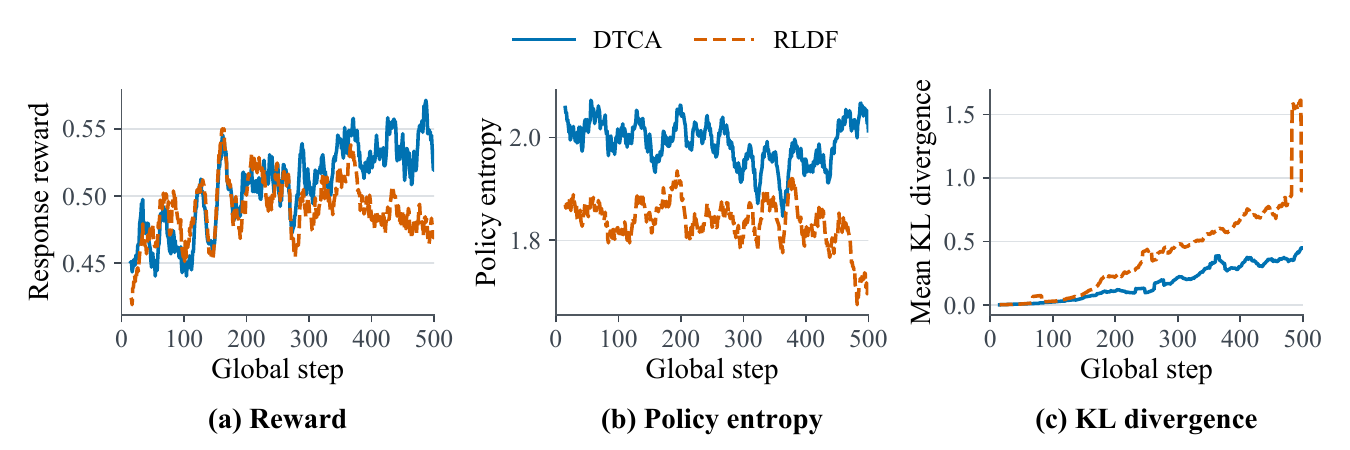}
    \caption{Additional entropy and training-dynamics analyses.
    Top: current-commit token entropy changes in DTCA without entropy
    calibration, using the same early-training baseline, 20-update bins,
    and color scale as Figure~\ref{fig:appendix_pos_policy_entropy_dtca}.
    Bottom: DTCA versus RLDF in (a) response reward, (b) policy entropy,
    and (c) mean KL divergence, using trailing 15-update means.}
    \label{fig:appendix_entropy_training}
    \label{fig:appendix_pos_policy_entropy_no_calibration}
    \label{fig:appendix_dtca_rldf_training}
\end{figure}

\paragraph{Training dynamics against RLDF.}
The lower panels of Figure~\ref{fig:appendix_entropy_training} compare the training dynamics of
DTCA and RLDF over 500 updates. The figure reports response reward, policy
entropy, and mean KL divergence, with curves smoothed by a trailing 15-update
mean. A higher response reward indicates that more sampled responses are
correct.

The two methods start with similar reward trajectories, with average rewards
of 0.464 for DTCA and 0.467 for RLDF over updates 1--100. Over updates
401--500, DTCA achieves a higher average reward of 0.537, compared with 0.482
for RLDF. During the same period, DTCA maintains higher policy entropy (2.00 vs.
1.78 nats) and lower mean KL divergence (0.36 vs. 0.89). These results show
that DTCA and RLDF follow different optimization trajectories, with DTCA
achieving higher reward while maintaining a smaller deviation from the
reference policy.

The observed differences are consistent with DTCA's design, where
contribution-based credit assignment provides token-dependent learning signals
and entropy-based calibration adjusts updates according to entropy changes.
This comparison reflects the combined effect of DTCA components and does not
isolate individual module contributions.

\begin{figure}[H]
    \centering
    \includegraphics[width=0.93\linewidth]{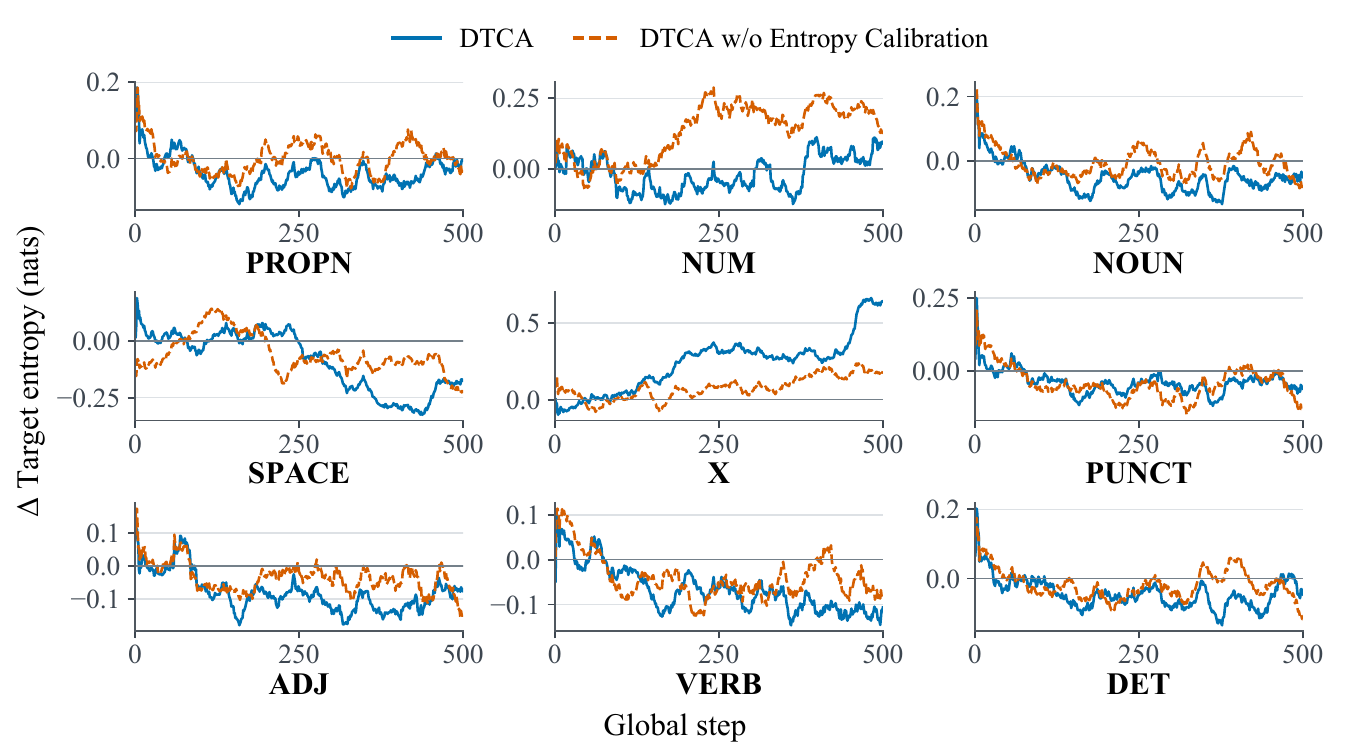}
    \caption{Predictive entropy changes for DTCA's nine most frequent target
    POS categories. Blue solid: DTCA. Orange dashed: without entropy calibration.
    Curves show trailing 25-update token-weighted means relative to each run's
    POS-specific baseline over updates 1--100.}
    \label{fig:appendix_pos_target_entropy_top9}
\end{figure}

\begin{figure}[H]
    \centering
    \includegraphics[width=0.93\linewidth]{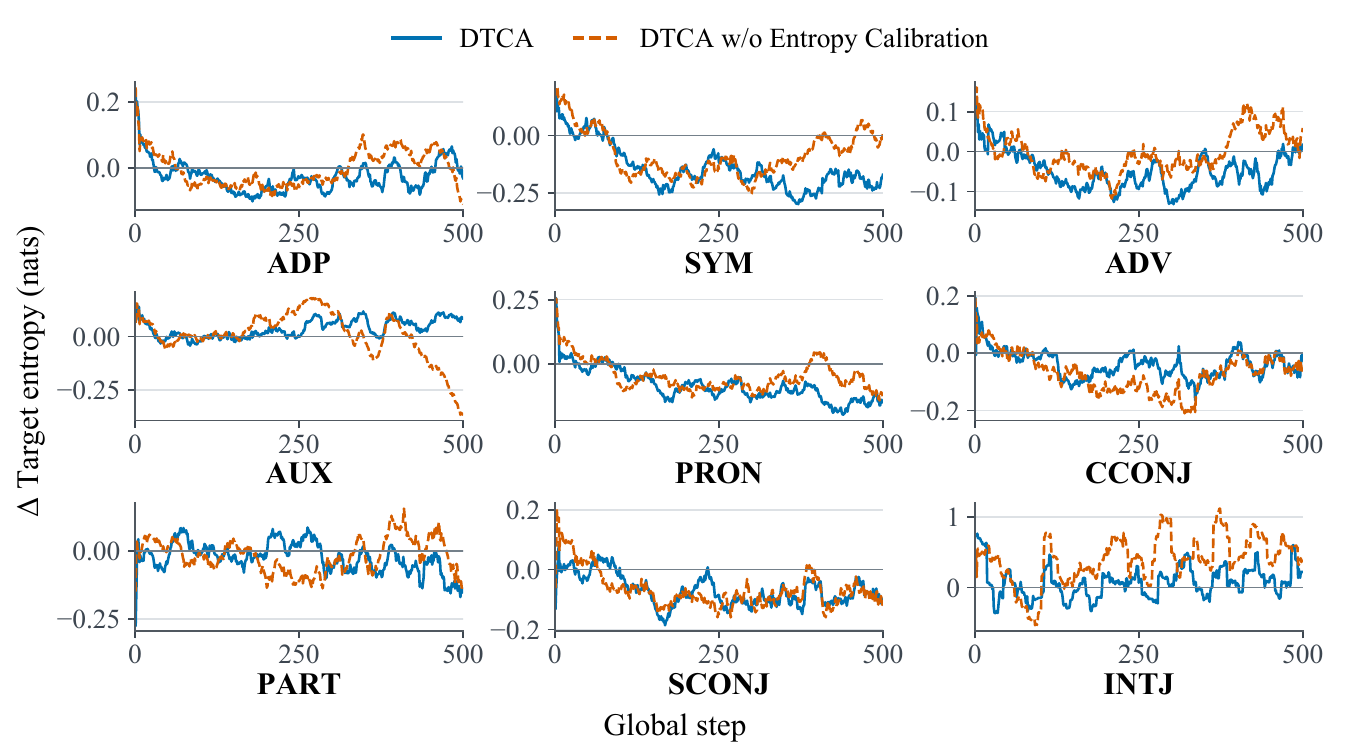}
    \caption{Predictive entropy changes for the nine remaining POS categories,
    ordered by their frequency among DTCA's retained targets. Colors, line
    styles, baseline, and smoothing match
    Figure~\ref{fig:appendix_pos_target_entropy_top9}.}
    \label{fig:appendix_pos_target_entropy_remaining9}
\end{figure}

\paragraph{Full POS entropy analysis.}
Figure~\ref{fig:appendix_pos_target_entropy_top9} shows the nine POS categories
most often retained as training targets in DTCA. Together with
Figure~\ref{fig:appendix_pos_target_entropy_remaining9}, it extends
Figure~\ref{fig:entropy_calibration_analysis} to all 18 POS categories.
These curves cover all retained targets at selected states, rather than only
current-commit tokens. With run-specific early baselines, positive values
indicate entropy increases, and curves compare changes rather than absolute
levels.

The frequent categories show different trends. Over updates
401--500, NUM is 0.051 nats above its early baseline in DTCA, compared with
0.209 nats without calibration. For X, the shifts are
0.447 and 0.184 nats. SPACE, ADJ, and VERB generally move below their early
baselines in both runs. Calibration does not uniformly lower entropy across POS.

The remaining categories also differ. Over updates 401--500, SYM falls by
0.192 nats from its early baseline in DTCA and by 0.014 nats without
calibration. AUX instead rises by 0.072 nats in DTCA and falls by 0.133 nats
without calibration. INTJ fluctuates sharply, warranting caution in
interpreting its trend.

\begin{figure}[H]
    \centering
    \includegraphics[width=\linewidth]{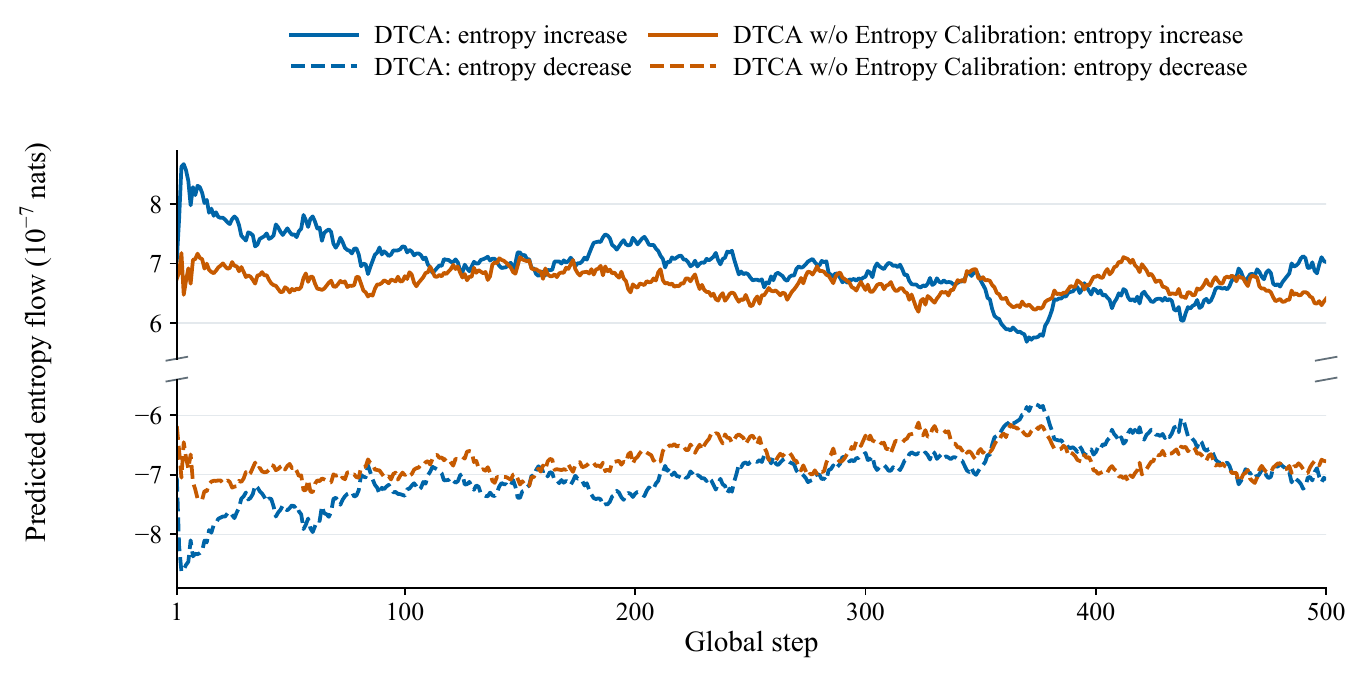}
    \caption{Predicted entropy-increasing (solid) and entropy-decreasing
    (dashed) components for DTCA (blue) and DTCA without entropy calibration
    (orange). Curves use trailing 25-update means. The vertical axis is broken,
    with values in $10^{-7}$ nats.}
    \label{fig:appendix_opefo_entropy_flow}
\end{figure}

\Needspace{7\baselineskip}
\paragraph{Entropy-flow calibration.}
Figures~\ref{fig:appendix_pos_target_entropy_top9} and
\ref{fig:appendix_pos_target_entropy_remaining9} summarize POS-level entropy
trajectories over training. These are within-run trends rather than changes
in matched tokens, since the models visit different states and retain
different targets. Figure~\ref{fig:appendix_opefo_entropy_flow} complements
these trajectories by examining the predicted entropy changes induced by
individual policy updates, separated into entropy-increasing and
entropy-decreasing parts.

Solid curves show the total predicted increase, while dashed curves show the
total predicted decrease. The broken vertical axis keeps both signs visible
despite their similar magnitudes. Equal magnitudes imply a near-zero net
predicted change.

DTCA keeps the two parts closer together. Over updates 1--100, its average
net predicted change is $-0.033\times10^{-7}$ nats, compared with
$-0.267\times10^{-7}$ nats without calibration. Over updates 401--500, the
corresponding values are $-0.081\times10^{-7}$ and $-0.201\times10^{-7}$
nats. Thus, the predicted increase and decrease nearly cancel under DTCA in
both windows. These values are first-order estimates based on the training
signals, not direct measurements of entropy after a parameter update.

\FloatBarrier
\Needspace{6\baselineskip}
\section{Limitations}
\label{app:limitations}

While DTCA provides an effective approach for fine-grained credit assignment in
diffusion language model reinforcement learning, several aspects remain open
for future exploration. DTCA uses POS categories to aggregate token-level
confidence improvements and obtain stable contribution estimates. This design
relies on the quality of POS alignment and annotation, and more accurate
token-level labeling strategies may further improve the estimation process.
Moreover, DTCA currently defines token contribution based on the confidence
improvement of remaining masked positions after token commitment. Developing
more principled theoretical formulations for contribution estimation in
diffusion language models remains an interesting direction. Finally, our
experiments focus on mathematical reasoning with LLaDA-8B-Instruct. Applying
DTCA to other tasks, such as code generation, may require adapting the token
grouping strategy to different token distributions and generation patterns.
Further evaluation on diverse diffusion language model architectures is also
needed to understand the general applicability of DTCA.

\end{document}